# Quantifying Interpretability and Trust in Machine Learning Systems

**Philipp Schmidt** and **Felix Biessmann**

Amazon Research

{phschmid, biessman}@amazon.de

**Abstract**

Decisions by Machine Learning (ML) models have become ubiquitous. Trusting these decisions requires understanding how algorithms take them. Hence interpretability methods for ML are an active focus of research. A central problem in this context is that both the quality of interpretability methods as well as trust in ML predictions are difficult to measure. Yet evaluations, comparisons and improvements of trust and interpretability require quantifiable measures. Here we propose a quantitative measure for the quality of interpretability methods. Based on that we derive a quantitative measure of trust in ML decisions. Building on previous work we propose to measure intuitive understanding of algorithmic decisions using the information transfer rate at which humans replicate ML model predictions. We provide empirical evidence from crowdsourcing experiments that the proposed metric robustly differentiates interpretability methods. The proposed metric also demonstrates the value of interpretability for ML assisted human decision making: in our experiments providing explanations more than doubled productivity in annotation tasks. However unbiased human judgement is critical for doctors, judges, policy makers and others. Here we derive a trust metric that identifies when human decisions are overly biased towards ML predictions. Our results complement existing qualitative work on trust and interpretability by quantifiable measures that can serve as objectives for further improving methods in this field of research.

## Introduction

In recent years machine learning (ML) models have shown to be competitive with human performance in some tasks. Predictions of ML models are used as assistive technology or even without human intervention in fully automated systems. When relying on decisions made by algorithms, humans need to trust these decisions. This is why *interpretability* of ML models and their decisions has become a major focus of research in recent years (Doshi-Velez and Kim 2017; Herman 2017).

One of the key problems with research on interpretability is that it is difficult for the community to agree on a definition (Lipton 2016). Consequently the quality of interpretability methods is difficult to measure and hence methods cannot be directly compared across studies in unified benchmark tests. Most of the research on interpretability compares methods using either proxy measures, that do not directly relate to interpretability by humans, as e.g. (Samek et al. 2017), or qualitative measures that render comparisons of results across studies difficult (Strumbelj and Kononenko 2010). In this work we propose a metric to quantify and compare the quality of interpretability methods.

Another equally challenging to measure concept that is tightly coupled to interpretability is *trust* in ML systems. Algorithmic decision making has huge value for improving human decision making and we provide empirical evidence for this in our study. Yet there are many areas in which we must ensure that human judgement is not dominated by algorithmic decisions. At court, in police stations or in hospitals we do not want decision makers to put too much trust in ML systems. We argue that it is of paramount importance to reliably measure when human decisions are overly biased to ML model predictions. Here we propose metrics to quantify trust in ML systems.

Lacking common metrics for interpretability quality and trust is not only problematic for practitioners that need to choose amongst an arsenal of different methods for interpretability, or that need to choose amongst a multitude of ML models the one that is most trustworthy. Even worse, without a commonly agreed upon objective to optimize, it is challenging for the research community as a whole to make progress in that space. The necessity of a gold standard test in order to enable us to argue on difficult to define philosophical concepts was the main motivation behind the famous Turing Test (Turing 1950): in the absence of proper definitions of *intelligence* it was difficult to argue about whether or not machines have it, so the author proposed a quantifiable metric that, if commonly agreed upon, can provide empirical evidence in an otherwise unstructured discussion. Along these lines of thought our work builds on existing interpretability research and extends it with the following contributions:

1. A quality score for interpretability methods

2. Empirical evidence demonstrating that interpretability can more than double audit productivity

3. Comparisons of popular interpretability approaches

4. A quantifiable trust score for ML models

This manuscript is structured as follows: after recapitulating some of the related work we will introduce a quality metric for evaluation and comparison of interpretability methods. Based on this metric we will derive a trust score for ML models. Thereafter we describe two text classification experiments we performed in order to demonstrate our approach, then we describe experimental results, we discuss some of their implications for evaluations of interpretability methods and we conclude with a summary of our findings.

## Related Work

The work on interpretability of ML models has become a central topic of research in both theoretical aspects of statistical learning as well as applied ML. Many of the relevant publications at major ML conferences and dedicated workshops can be broadly categorized in more conceptual contributions or position papers and technical contributions to the field of interpretability.

In the category of position papers, an important aspect dealt with in (Herman 2017) is the question of how we balance our concerns for transparency and ethics with our desire for interpretability. Herman points out the dilemma in interpretability research: there is a tradeoff between explaining a model's decision *faithfully* and *in a way that humans easily understand*. Interpreting ML decisions in an accessible manner for humans is also referred to as *simulatability* (Lipton 2016). Our work builds on these findings in that we leverage this aspect of intuitive accessibility or cognitive friction of ML prediction explanations as the basis of our metric. In a similar vein the authors of (Doshi-Velez and Kim 2017) highlight the necessity of *understandability* of explanations as well as the lack of consensus when it comes to evaluating interpretability of ML models. They propose an evaluation taxonomy that comprises both automated evaluations but also involves evaluations by human laymen. In summary there appears to be a consensus in the conceptual work on interpretability that a) good model explanations overlap with human intuitions and b) there is a lack of quantitative evaluation standards (Miller 2017; Doshi-Velez and Kim 2017; Lipton 2017).

The category of technical contributions can be further subdivided into two types of methods. First there are methods that aim at rendering specific models interpretable, such as interpretability methods for linear models (Haufe et al. 2014) or interpretability for neural network models (Zeiler and Fergus 2014; Simonyan, Vedaldi, and Zisserman 2013; Montavon et al. 2017). Second there are interpretability approaches that aim at rendering *any* model interpretable, a popular example are the *Local Interpretable Model-Agnostic Explanations* (LIME) (Ribeiro, Singh, and Guestrin 2016). As these latter interpretability methods work without considering the inner workings of an ML model, they are often referred to as *black box interpretability methods*. We will use one method from each of these two categories, black box and glass box models, in our experiments. Note however that the focus of this work is not to advocate a specific interpretability method, but rather an evaluation strategy for such methods. We thus focus in the following on the evaluation aspect of related work.

The most straightforward approach to evaluation of interpretability is to generate synthetic data from a known generative model and evaluate the explanations against the true data generation process, examples are for instance (Zien et al. 2009; Haufe et al. 2014). As appealing this approach might be, it can be very challenging to design generative models for real data. Recovering generative models on synthetic data is an important sanity check for any interpretability method, but for practically relevant tasks, we need to be able to compare explanations for models trained on real world data sets.

Due to the popularity of neural network models especially in the field of computer vision there have been a number of interpretability approaches specialized for that application scenario and the method of choice in this field, deep neural networks. Some prominent examples are *layerwise relevance propagation* (LRP), sensitivity analysis (Simonyan, Vedaldi, and Zisserman 2013) and deconvolutions (Zeiler and Fergus 2014). For comparing these different approaches the authors of (Samek et al. 2017) propose a greedy iterative perturbation procedure for comparing LRP, sensitivity analysis and deconvolutions. The idea is to remove features where the perturbation probability is proportional to the relevance score of each feature given by the respective interpretability method. An interesting finding in that study is that the results of interpretability comparisons can be very different depending on the metric: the authors of (Goodfellow, Shlens, and Szegedy 2014) performed an evaluation of sensitivity analysis and came to a different conclusion than (Samek et al. 2017).

The idea of using perturbations gave rise to many other interpretability approaches, such as the work on *influence functions* (Cook 1977; Koh and Liang 2017; Hampel et al. 2011) and methods based on game theoretic insights (Strumbelj and Kononenko 2010; Lundberg and Lee 2017). In (Strumbelj and Kononenko 2010) evaluations are entirely qualitative; in (Lundberg and Lee 2017) the authors compare interpretability methods by testing the overlap of explanations with human intuitions. While this approach can be considered quantitative, it is difficult to scale as it requires task specific user studies. Another metric used in that study for comparisons of evaluations is computational efficiency, which is simple to quantify, but is not directly related to interpretability. Other studies also employ user studies and comparisons with human judgements of feature importance. An interesting approach is taken in (Ribeiro, Singh, and Guestrin 2018) in which the authors let students of an ML class guess what a model would have predicted for a given instance when provided with an explanation. Similarly the authors of (Lakkaraju, Bach, and Leskovec 2016) perform user studies in which they present rules generated by an interpretability method and measure how good and how fast students can replicate model predictions. This approach has been taken also in (Huysmans et al. 2011) in which the speed and accuracy are measured with which humans can replicate model decisions. Both of these studies were conducted in a relatively controlled environment and with skilled subjects. Also the explanations provided in these user studies were rather complex which means that the explanations are mod-el specific and cannot be parsed intuitively by laymen. In contrast to this work we consider a crowdsourcing scenario that allows to scale evaluations leveraging the workforce of laymen.

Our contribution is based on the above findings that relate the quality of interpretability models to human intuitions, however with some important differences. While some evaluation strategies of interpretability methods are entirely qualitative, e.g. (Strumbelj and Kononenko 2010), our approach is quantitative. Moreover our approach relates directly to interpretability and not to computational efficiency, as in (Lundberg and Lee 2017), or to robustness of the model under perturbations as in (Samek et al. 2017). Another advantage of the proposed evaluation is that it is scalable as it does not require prior knowledge or expertise nor does it require task specific preparation as in (Letham et al. 2015; Lakkaraju, Bach, and Leskovec 2016); instead our proposed evaluation can be performed in the background of a standard audit procedure of a ML task in a production setting. Importantly we show that when using appropriate interpretability methods our approach can more than double efficiency in crowdsourcing tasks.

## Measuring Quality of ML Explanations

In order to quantify the quality of interpretability methods we consider the following three propositions:

**Proposition 1** *The quality of an interpretability method should be independent of the task and the ML method used to solve the task.*

The outcome and quality of interpretability methods usually depend on a particular task and ML model. Some tasks and models are simpler to interpret than others, but an interpretability method should be invariant with respect to those factors. In order to quantitatively compare interpretability methods both task and model should be kept constant.

**Proposition 2** *The quality of an interpretability approach should capture* intuitive understanding.

Most definitions of interpretability rely on some notion of intuitive understanding. Explanations of a model's decision are better if they are comprehensible. As we do not intend to replace the semantic ambiguity of *interpretability* with another one we leverage the fact that *intuitive understanding* can be measured:

**Proposition 3** *Faster and more accurate decisions indicate intuitive understanding.*

Based on the above statements we propose a metric that is based on the *increase in information transfer rate* in an annotation task when model explanations are provided. The idea is that the better an interpretability method, the faster and more accurately annotators will reproduce decisions of a given ML model. The metric builds on existing work that quantifies response times in user studies as well as agreement between annotators and model predictions (Huysmans et al. 2011; Lakkaraju, Bach, and Leskovec 2016). In contrast to these studies we propose to integrate these two notions that captures intuitive understanding, speed and agreement of predictions, into one measure that can be compared across a variety of models, tasks and interpretability approaches. The information transfer rate *ITR* can be measured in bit per second as

$$\text{ITR} = \frac{I(\hat{\mathcal{Y}}_\text{H}, \hat{\mathcal{Y}}_\text{ML})}{t} \quad (1)$$

where $t$ is the average response time in an annotation task and $I(\hat{\mathcal{Y}}_\text{H}, \hat{\mathcal{Y}}_\text{ML})$ denotes the mutual information between $\hat{\mathcal{Y}}_\text{H}$, the annotations provided by human labelers, and the model predictions $\hat{\mathcal{Y}}_\text{ML}$. For a given training data set and ML model, different interpretability methods will return different explanations. For each interpretability method, we measure the ITR when no explanation is provided during the annotation task and the ITR when annotators are provided with an explanation of what a ML model would have predicted. The actual model decisions are not shown to the annotators. We then compute the ITR increase with explanations in order to compare interpretability methods. Note that the ML model is optimized for the task of that data set but it was trained on a held out data set that the annotators do not see. The optimal model is typically as good as or better than humans at the task (for details see section *Results*).

The numerator of Equation 1 denotes the mutual information $I(\hat{\mathcal{Y}}_\text{ML}, \hat{\mathcal{Y}}_\text{H})$ between the model predictions and the labels obtained in the auditing task which can be computed as

$$I(\hat{\mathcal{Y}}_\text{ML}, \hat{\mathcal{Y}}_\text{H}) = \sum_{\hat{y}_\text{ML}, \hat{y}_\text{H}} p(\hat{y}_\text{ML}, \hat{y}_\text{H}) \log \frac{p(\hat{y}_\text{ML}, \hat{y}_\text{H})}{p(\hat{y}_\text{ML}) p(\hat{y}_\text{H})}, \quad (2)$$

where $\hat{y}_\text{ML}$ refers to the model predictions and $\hat{y}_\text{H}$ refers to the values provided by the human annotator. If the logarithm in Equation 2 is taken with respect to the basis of 2, then the resulting mutual information is measured in bit. This allows for easier comparisons of the ITR across tasks and models. To capture the temporal aspect of intuitive understanding we record the time it took a human labeler to provide an annotation with or without explanations. The longer an annotation task takes, the lower the ITR becomes. If explanations help annotators to understand model decisions, then annotation time will decrease and the ITR increases. If the time it takes an annotator to provide a label is measured in seconds, the ITR is thus measured in bit per second.

## Measuring Trust in ML Models

As interpretability and trust in a ML system are intimately related, we propose a novel *trust metric* based on the above interpretability metric. One could argue that interpretability in the sense of *replicability* is already a good measure of trust. However that definition does not account for the quality of the model predictions. A trust metric must capture cases in which humans are too biased towards the decisions of a ML system and overly trust the system. In other words if model predictions are wrong and humans still agree with model predictions, the coefficient should be large. In contrast if humans do not agree with the model predictions when they are wrong, or if humans are sceptical and take longer to make a decision in those often difficult cases

then the trust coefficient should be low. More formally we propose to measure trust in a ML system as

$$\bar{T} = \frac{\text{ITR}_{\hat{y}_{\text{ML}}}}{\text{ITR}_{\mathcal{Y}}} \quad (3)$$

where $\text{ITR}_{\hat{y}_{\text{ML}}}$ denotes the ITR as defined in Equation 1 measured using the mutual information between human decisions and model predictions and $\text{ITR}_{\mathcal{Y}}$ is measured using the mutual information between human decisions and true labels. This coefficient will be larger than one if humans are more biased towards the predictions of a ML system compared to the true labels of a task. The coefficient will be smaller than one if humans rely more on their own judgement. Note that $\bar{T}$ not only captures label agreement but also how much time is spent for decision making. We believe that this is a crucial factor when examining trust in ML systems.

## Experiments

We ran two different annotation experiments for evaluating the quality of interpretability methods. The following paragraphs describe the data sets, the machine learning models, the interpretability approaches used in the experiments and the experimental paradigm.

### Data Sets

We chose two text classification data sets for our annotation experiments. The first one is a book categorization task and the second one is a binary sentiment classification task of IMDb reviews.

**Book Categories** We have used a proprietary dataset on book categories. The dataset contains $744,463$ books in nine categories. We included only books where the text length of the product description was more than $200$ characters long and we randomly split the data set into a train and test set with a $30\%/70\%$ ratio. The class counts in the test data set used for the annotation task were between 500 and 1000 data points and the performance as measured by class-frequency weighted precision/recall/f1 of the ML model on a held out test data set for this task is above $85\%$, for details see Table 1.

**IMDb** In another experiment we used the publicly available IMDb movie review sentiment dataset [1] which was introduced in (Maas et al. 2011). The IMDb rating scale is defined from one to ten stars where all reviews that have less than five stars are considered to have negative sentiment and all reviews that have more than six stars positive. For the prepared dataset we have sampled 128 movie reviews, all between 400 and 1000 characters long, for each star rating from the test set resulting in a total of 1024 movie reviews. The ML model was trained on the complete train dataset which consists of 25000 movie reviews and achieved precision/recall/f1 metrics close to $90\%$ on the test dataset, for details see Table 2.

---

[1] https://www.imdb.com/conditions

| Category | precision | recall | f1-score | support |
|---|---|---|---|---|
| 1 | 0.76 | 0.86 | 0.81 | 620 |
| 2 | 0.89 | 0.78 | 0.84 | 573 |
| 3 | 0.81 | 0.72 | 0.76 | 609 |
| 4 | 0.92 | 0.88 | 0.90 | 937 |
| 5 | 0.80 | 0.92 | 0.86 | 886 |
| 6 | 0.78 | 0.80 | 0.79 | 649 |
| 7 | 0.88 | 0.82 | 0.85 | 585 |
| 8 | 0.99 | 0.98 | 0.99 | 590 |
| 9 | 0.96 | 0.96 | 0.96 | 681 |
| avg / total | 0.87 | 0.86 | 0.86 | 6130 |

Table 1: Held-out per label precision/recall/f1 scores of the ML model used for comparing ML interpretability methods on the book category dataset. Class-frequency weighted metrics are above $85\%$ which can be considered on par with or better than human annotator performance, see also *Results* section.

| sentiment | precision | recall | f1-score | support |
|---|---|---|---|---|
| negative | 0.88 | 0.87 | 0.87 | 12500 |
| positive | 0.87 | 0.88 | 0.87 | 12500 |
| avg / total | 0.87 | 0.87 | 0.87 | 25000 |

Table 2: Held-out per label precision/recall/f1 scores of the ML model used for comparing ML interpretability methods on the IMDb dataset.

### Machine Learning Model

In all experiments we used unigram bag-of-words features that were term-frequency inverse document frequency normalized. English stopwords were removed prior to the feature extraction. Bag-of-words feature vectors $\mathbf{x} \in \mathbb{R}^d$, where $d$ denotes the number of unigram features, were used to train an $L_2$ regularized multinomial logistic regression model. Let $y \in \{1, 2, \ldots, K\}$ be the true label, where $K$ is the total number of labels and $\mathbf{W} = [\mathbf{w}_1, \ldots, \mathbf{w}_K] \in \mathbb{R}^{d \times K}$ is the concatenation of the weight vectors $\mathbf{w}_k \in \mathbb{R}^d$ associated with the $k$th class then

$$p(y = k | \mathbf{x}, \mathbf{W}) = \frac{e^{z_k}}{\sum_{j=1}^{K} e^{z_j}} \quad \text{with } z_k = \mathbf{w}_k^\top \mathbf{x} \quad (4)$$

We estimated $\mathbf{W}$ using stochastic gradient descent (SGD) using the python library `sklearn` and used a regularization parameter of $0.0001$, other values for the regularizer lead to similar model performance.

### Interpretability Methods

We chose two different interpretability methods for providing explanations in our annotation tasks, *Local Interpretable Model-Agnostic Explanations* (LIME) (Ribeiro, Singh, and Guestrin 2016) and a simple univariate measure of covariance between features and model predictions (Haufe et al. 2014), referred to as COVAR in the following. The selection of these two was motivated by their general applicabil-ity, by their popularity (in the case of LIME) and by their simplicity and speed (in the case of COVAR). Both methods can be considered as generic enough to cover a variety of ML models and application scenarios, even if we only apply them to text classification tasks in this work. LIME can be applied to any ML model and the covariance based method can be applied to most linear models, however there are also extensions for non-linear models (Zien et al. 2009; Kindermans et al. 2017).

**LIME** LIME (Ribeiro, Singh, and Guestrin 2016) belongs to the model agnostic family of *black-box* interpretability methods. Given a pretrained model and a data point, LIME will perturb the data point and compute a model prediction for each perturbation. Perturbations are artificial new data points where random words are dropped from the input text. In our case, for computing the top-3 most relevant uni-grams, LIME is using forward feature selection and ranks the uni-grams according to the coefficients of the local linear model.

**COVAR** Similar to the local linear approximations of LIME one can consider a global interpretable linear model. The implicit assumption when interpreting coefficients of linear models is that the features of the training data were uncorrelated – which is rarely the case in real data sets[2]. A simple and efficient approach for rendering linear models interpretable is provided in (Haufe et al. 2014). While this approach is limited to linear models, it is a special case of the feature importance ranking measure (FIRM) (Zien et al. 2009) that can be applied to arbitrary non-linear models and there are other non-linear extensions (Kindermans et al. 2017). Following Eq. (6) in (Haufe et al. 2014) we obtain the feature importances of the COVAR approach for each class $k$ separately by

$$\mathbf{a}_k^{\text{COVAR}} = \mathbf{X}^\top \hat{\mathbf{y}}_k \quad (5)$$

where the matrix $\mathbf{X} \in \mathbb{R}^{N \times d}$ denotes the N samples in the held out test data set and the and $d$ denotes the number of unigram features extracted from the training data set. The predictions of the model for class $k$ on the test data are denoted $\hat{\mathbf{y}}_k \in \mathbb{R}^{N \times 1}$. Each dimension of $\mathbf{a}_k^{\text{COVAR}} \in \mathbb{R}^d$ is associated with a feature, in our case a word in the unigram dictionary. To compute the explanations, i.e. the highlighted words for sample $\mathbf{x}_i$, we selected the feature importances $\mathbf{a}_k$ associated with the most likely predicted class $k$ under the model and ranked the words in a text according to the element-wise product of features $\mathbf{x}_i$ and feature/prediction covariances $\mathbf{a}_k$. The highlighted words were those that were present in the text and scored high in terms of their covariance between features and model predictions.

## Experimental Setup

All user study experiments were run on Mechanical Turk where we asked annotators to provide the correct produc-t category for a given product description of a book or the sentiment of a movie review. The user interface (UI), shown in Figure 1, and the data collection scheme were the same for both experiments. For each data point in a data set we collected nine annotations from distinct workers. For each annotation assignment we selected one out of three conditions uniformly at random to apply to the to-be-labelled texts. The conditions were a) showing the text without explanations (no-highlights), b) showing text with words highlighted according to the LIME explanations or c) showing texts with words highlighted according to COVAR explanations. To select the top-3 most important words according to LIME we used `LimeTextExplainer.explain_instance()` with 2500 pertubed samples. To control for effects related to the number of words highlighted we kept the number of words highlighted fixed to three words in each text, samples with more words highlighted (e.g. due to duplicate words) were discarded. Similarly the length of the texts were controlled for. For each annotation we recorded the annotation time, the experimental condition, the true label and the label provided by the annotator. As additional sanity check we conducted a separate experiment in which we showed randomly highlighted words for the book categorization task.

---

[2] As the perturbations of the LIME approach usually result in uncorrelated features, the linear model parameters of the LIME model approximations can be interpreted directly.

Figure 1: Example of an annotation task UI showing an example for positive sentiment from the IMDb dataset with COVAR highlights. Words with high scores of an interpretability approach are highlighted in yellow. To ensure accurate time measurements we only showed the text for the task and started the task timer after the start button at the top of the page was clicked.

## Results

In this section we compare the annotation accuracy and annotation speed of the workers when presented with just the text and when presented with the text *and* an explanation of what a model would have predicted. Note that the ML model predictions were not shown to the annotators and that the ML model was invisible to the workers. The data sets as well as the ML model were fixed for each of the conditions compared in the following. All results related to the performance of the ML models are described in the previous section and are not in the focus of the following comparisons.

| Class | no-highlights | | LIME | | COVAR | |
|---|---|---|---|---|---|---|
| | acc | time | acc | time | acc | time |
| 1 | 65.19 | **8.64** | 66.75 | 8.86 | **68.95** | 9.83 |
| 2 | 51.20 | 9.26 | 59.66 | 8.87 | **62.98** | **8.61** |
| 3 | 44.09 | 8.85 | 45.72 | 8.92 | **46.98** | **8.76** |
| 4 | 62.61 | 7.06 | **68.82** | 6.46 | 68.49 | **6.08** |
| 5 | 86.62 | 9.16 | 87.37 | 8.36 | **89.03** | **7.88** |
| 6 | 40.47 | 9.69 | 42.09 | **8.54** | **42.76** | 8.85 |
| 7 | 39.87 | 11.44 | 49.22 | 9.35 | **54.90** | **9.24** |
| 8 | 87.32 | 6.91 | 89.55 | 5.85 | **92.86** | **5.67** |
| 9 | 85.33 | 7.72 | 83.75 | **6.98** | **88.73** | 7.76 |
| avg | 61.67 | 8.67 | 64.50 | **8.10** | **65.87** | 8.12 |

Table 3: Experimental results for the book category dataset showing average accuracies (*acc*) in percent and task times (*time*) in seconds by true book category (*Class*) and applied condition for the HIT. Explanations improve annotator accuracy in all cases with the simpler COVAR explanations resulting in slightly higher accuracies than LIME explanations. In all but one ambiguous class showing annotators explanations of ML predictions yielded a decreased annotation time compared to the control condition. Bottom row lists class frequency weighted averages.

**Results on Book Categories**

We show the annotation accuracy and task times for the book categories data set in Table 3. In all classes we observed an improvement in annotation accuracy when presenting explanations. On average an absolute accuracy uplift of more than 4% was observed compared to the control condition without explanations. We found the COVAR condition to result in slightly higher annotator accuracies compared to LIME. Annotators' response times were decreased in all but one class which happened to be a rather ambiguous category. On average we measured a relative speedup of more than 6%. Comparing the different explanations we saw a larger speedup with the COVAR approach in six out of nine cases. Both increased accuracy and decreased annotation time were statistically significant (chi-square test of independence, $p < 0.01$ and Kruskal-Wallis test for annotation times, $p < 0.01$).

**Results on IMDb sentiment**

The results from the IMDb sentiment task are similar to the book categorization task. Providing explanations in annotation tasks consistently increases the annotation accuracy and decreases the annotation response time. In Table 4 we show that annotators can improve their accuracy of 78.01% in the control condition to an accuracy of 81.72% with LIME explanations and to an accuracy of 84.52% with the COVAR explanations. The average decrease in annotation time is even more pronounced with a reduction to 7.18s (LIME) and 6.66s (COVAR) relative to 8.91s in the control condition. Also in this task the increase in accuracy and the decrease in annotation time was statistically significant (chi-square test of independence, $p < 0.01$ and Kruskal-Wallis test for annotation times, $p < 0.01$).

| Sent. | no-highlights | | LIME | | COVAR | |
|---|---|---|---|---|---|---|
| | acc | time | acc | time | acc | time |
| neg. | 68.43 | 9.54 | 72.51 | 7.82 | **76.77** | **7.07** |
| pos. | 87.20 | 8.30 | 91.17 | 6.53 | **92.43** | **6.25** |
| avg | 78.01 | 8.91 | 81.72 | 7.18 | **84.52** | **6.66** |

Table 4: Experimental results for the IMDb dataset showing average accuracies (*acc*) in percent and task times (*time*) in seconds by true sentiment label (*Sent.*) and applied condition for the HIT. Overall, a relative speedup of more than 33% was observed when applying COVAR compared to providing no explanation to annotators. Also showing explanations substantially improved classification accuracy in all cases. Bottom row lists class frequency weighted averages.

| Mutual Inform. | Task | Information Transfer Rate (bit/s) | | |
|---|---|---|---|---|
| | | no-highlights | LIME | COVAR |
| $\hat{\mathcal{Y}}_{ML}$ | **Books** | 0.156 | 0.179 | **0.189** |
| | **IMDb** | 0.020 | 0.045 | **0.072** |
| $\mathcal{Y}$ | **Books** | 0.165 | 0.187 | **0.197** |
| | **IMDb** | 0.028 | 0.047 | **0.060** |

Table 5: *Top:* The quality of two interpretability methods LIME and COVAR, measured as Information Transfer Rate (ITR, see Equation 1) in bit/s between annotations by human labelers $\hat{\mathcal{Y}}_H$ and ML model predictions $\hat{\mathcal{Y}}_{ML}$. With COVAR explanations annotators have a 21% (Books) or 260% (IMDb) higher ITR compared to annotations where no explanations were provided. *Bottom:* Also ITRs between annotations $\hat{\mathcal{Y}}_H$ and true labels $\mathcal{Y}$ show significant improvements, 19% and 113% for Books and IMDb respectively, when COVAR model explanations are shown.

**Comparing Interpretability Approaches**

Based on the reasoning in propositions 1 to 3 we use the proposed measure of information transfer rate (ITR) to compare the quality of interpretability methods. The ITRs listed in Table 5 show that both interpretability methods examined help auditors to gain intuitive understanding of a model's predictions. As additional sanity check we compared the ITR with conditions in which annotators were shown randomly highlighted words. Those results were obtained in a separate experiment and only for the book categorization. The ITR in that condition was reduced by 4.5% compared to the condition when no explanations were shown. Across all conditions we find that auditors find it easier to understand model predictions with the COVAR approach (see Equation 5). This is an interesting finding, as the COVAR approach is computationally more efficient than LIME. We measured wallclock running times for computing the explanations for one data point with each of the interpretability methods, LIME and COVAR. We ran 64 repeated measure-ments on a MacBook Pro 2017 equipped with a 2,5 GHz Intel Core i7. Computing a LIME explanation for one test instance with 2500 pertubations took $0.750 \pm 0.111$ seconds (mean/standard deviation) whereas computing the top pattern activations for COVAR took $0.014 \pm 0.008$ seconds. On average computing explanations with COVAR is 53 times faster than with LIME.

| Task  | Trust coefficients $\bar{T}$ | | | |
|-------|--------|-------------|-------|-------|
|       | Random | no-highlights | LIME | COVAR |
| **Books** | 0.673 | 0.946 | 0.956 | 0.959 |
| **IMDb**  | –     | 0.713 | 0.961 | 1.215 |

Table 6: Trust coefficients $\bar{T}$ (see Equation 3) for all evaluated conditions. The COVAR trust coefficient is above 1 for the IMDb data-set, reflecting a bias of annotators towards model predictions; this indicates that annotators overtrust the model when COVAR explanations are provided. Explanations by LIME and no highlights resulted in trust coefficients close to or lower than 1. Randomly highlighted words yielded the lowest trust coefficient.

**Explanations improve audit efficiency**

The ITR between human annotations $\hat{\mathcal{Y}}_H$ and model predictions $\hat{\mathcal{Y}}_{ML}$ (top row in Table 5) measures intuitive understanding of model decisions and hence serves as marker for model interpretability. We also investigated $I(\hat{\mathcal{Y}}_H, \mathcal{Y})$, the ITR between human annotations $\hat{\mathcal{Y}}_H$ and the *true* labels $\mathcal{Y}$. This quantity measures the value of interpretability methods for improving the productivity in an audit task, or more generally interaction between humans and AI systems. The results in Table 5, bottom row, demonstrate that the performance of auditors in annotation tasks improves significantly when explanations of ML model predictions were provided. Without any explanations, the ITR in the books and IMBd task were 0.165 and 0.028, respectively; with explanations annotators achieved an ITR of up to 0.197 (books) and 0.06 (IMDb), which amounts to a 19% improvement in the books task and a 113% improvement in the IMDb task. This means that providing explanations of ML models that are otherwise invisible to the auditors, can lead to a more than *twofold* increase in audit productivity.

**Quantifying Trust in ML Predictions**

We next examined the extent to which humans trust ML model predictions using the *trust coefficient* $\bar{T}$ in Equation 3. The results are shown in Table 6. A first important finding is that when no explanations or random highlights are shown, humans do not tend to replicate the mistakes made by a model, as indicated by the trust coefficients below one in both tasks. This holds true also for the condition in which LIME explanations are shown. The most interesting finding is in the condition in which COVAR explanations were shown. While in the book categorization task the trust coefficient below one indicates that humans still follow their own judgement, the trust coefficient larger than one in the IMDb task suggests that humans were biased towards the ML predictions even if the predictions were wrong.

## Conclusion

Understanding algorithmic decisions has been driving research on interpretability methods for years but there was no gold standard strategy for evaluating these methods. In this work we propose metrics that can be used as unified quality measure of interpretability. The information transfer rate at which humans understand and replicate model predictions, can be measured in bit/s and thus allows for easier comparison across tasks, models and interpretability approaches. When comparing two popular interpretability approaches, COVAR, a glass-box method, and LIME, a black-box method with respect to the ITR, we find that COVAR yields more interpretable explanations. This highlights the potential of simple methods like COVAR for some combinations of data sets and ML models. Note however that we do advocate a particular interpretability method, our goal was to provide a metric to evaluate and compare interpretability approaches for a given data set and ML model. We hope that the proposed metric will help practitioners to choose an interpretability method and that it will help researchers to improve existing approaches to understanding AI systems.

Beyond the evaluation of interpretability our experiments also provide empirical evidence for how interpretability can improve interaction between humans and AI systems: when provided with explanations auditors can more than double their productivity in an annotation task. We are convinced that this finding in combination with the proposed metric can serve as a basis for improving AI as assistive technology with the help of interpretability methods.

Building on these findings we derived a measure of trust in ML systems that captures whether humans are overly biased towards ML model predictions. Our results demonstrate that the proposed measure identifies cases in which human decisions deviate from their own judgement and are more reflective of what a ML system would have predicted. We believe that assessing these effects is crucial for drawing the ethical and legal boundaries of algorithmic decision making. As shown in our experiments ML systems are tremendously helpful for improving human decision making; but we argue that there are many examples in which human judgement must not be replaced with or dominated by ML systems. We hope that our work will be useful for policy makers that need to measure bias in human judgement in order to ensure human decisions will be improved by but not replaced by algorithmic decision making.

## Acknowledgments

We thank Tammo Rukat and Sebastian Schelter for helpful comments on the manuscript.